\documentclass[conference]{IEEEtran}
\IEEEoverridecommandlockouts
\usepackage{cite}
\usepackage{amsmath,amssymb,amsfonts}
\usepackage{algorithmic}
\usepackage{graphicx}
\usepackage{textcomp}

\usepackage{amsfonts}
\usepackage{booktabs}
\usepackage{multirow}
\usepackage{makecell}
\usepackage{bbding}
\usepackage[dvipsnames,table,cmyk]{xcolor} 
\usepackage{url}
\usepackage{listings}

\usepackage{tcolorbox}
\usepackage{threeparttable}

\def\BibTeX{{\rm B\kern-.05em{\sc i\kern-.025em b}\kern-.08em
    T\kern-.1667em\lower.7ex\hbox{E}\kern-.125emX}}
\begin{document}

\title{GiVE: Guiding Visual Encoder to Perceive Overlooked Information}

\DeclareRobustCommand*{\IEEEauthorrefmark}[1]{%
    \raisebox{0pt}[0pt][0pt]{\textsuperscript{\footnotesize\ensuremath{#1}}}}
\author{\IEEEauthorblockN{Junjie Li\IEEEauthorrefmark{1},
Jianghong Ma\IEEEauthorrefmark{1,2}*,
Xiaofeng Zhang\IEEEauthorrefmark{1}*, 
Yuhang Li\IEEEauthorrefmark{1},
and Jianyang Shi\IEEEauthorrefmark{1}
\thanks{* denotes corresponding authors. 
This work was partially supported by the National Natural Science Foundation of China (Project No. 62202122 and No. 62073272), the Guangdong Basic and Applied Basic Research Foundation under Grant No. 2024A1515011949, and the Shenzhen Fundamental Research Program under No. JCYJ20240813104837050 and No. GXWD20231130110308001.
}
}
\IEEEauthorblockA{\IEEEauthorrefmark{1}Harbin Institute of Technology, Shenzhen, China}
\IEEEauthorblockA{\IEEEauthorrefmark{2}City University of Hong Kong, China}
\IEEEauthorblockA{\{22b351018, 22s151185, 19b951026\}@stu.hit.edu.cn,  \{zhangxiaofeng, majianghong\}@hit.edu.cn}
}

\maketitle

\begin{abstract}
Multimodal Large Language Models have advanced AI in applications like text-to-video generation and visual question answering. These models rely on visual encoders to convert non-text data into vectors, but current encoders either lack semantic alignment or overlook non-salient objects. We propose the Guiding Visual Encoder to Perceive Overlooked Information (GiVE) approach. GiVE enhances visual representation with an Attention-Guided Adapter (AG-Adapter) module and an Object-focused Visual Semantic Learning module. These incorporate three novel loss terms: Object-focused Image-Text Contrast (OITC) loss, Object-focused Image-Image Contrast (OIIC) loss, and Object-focused Image Discrimination (OID) loss, improving object consideration, retrieval accuracy, and comprehensiveness. Our contributions include dynamic visual focus adjustment, novel loss functions to enhance object retrieval, and the Multi-Object Instruction (MOInst) dataset. Experiments show our approach achieves state-of-the-art performance. 
\end{abstract}


\vspace{-1mm}
\section{Introduction}
\label{sec:intro}
\vspace{-1mm}
Multimodal Large Language Models (MLLMs) \cite{blip-2} have advanced general artificial intelligence with strong generation and inference capabilities in applications such as text-to-video generation \cite{videoldm}, visual question answering \cite{Amanpreet_VQA}, and embodied robotics \cite{zhou2025pa}. A common architecture combines a visual encoder with a Large Language Model (LLM), embedding non-textual data into vectors interpretable by the LLM via a mapping mechanism. While research \cite{llava-1.5} highlights the effectiveness of this design, the quality of image embeddings from the visual encoder remains critical to MLLM performance.

An image encoder is a specialized visual encoder designed to map high-dimensional image data to a lower-dimensional feature space. These encoders can be broadly categorized based on their pre-training tasks into two main types: reconstruction-based and cross-modal contrastive learning-based encoders. 
Image encoding models trained with \textbf{reconstruction tasks} \cite{vqvae,mirl} are proficient in capturing comprehensive image details.  However, these models \textit{lack semantic alignment with text} during training, which complicates the LLM's ability to interpret image embeddings \cite{survey_mllm, zhou2023fcboost}. Consequently, such encoders are infrequently utilized in MLLMs. 
Vision Transformer (ViT) models trained with \textbf{image-text contrastive learning} \cite{clip,zhou2023bc} generally \textit{align effectively with LLMs} but face an implicit ``ignore'' problem, limiting their expressive capability. This limitation arises because different modalities convey distinct types of information. For instance, an image may feature multiple objects with unique attributes, such as texture, color, spatial location, and potential interactions. In contrast, \textbf{abstract text} typically highlights only \textbf{the most salient objects} and provides limited descriptions of other visual elements. ViTs trained for image-text matching tend to focus on the salient regions of the image that correspond to the text, thereby \textbf{overlooking secondary elements} like the background. MLLMs using such visual encoders exhibit diminished response quality when users inquire about non-salient objects.
In summary, for effective integration with LLMs, MLLMs require an image encoder that is both (1) semantically aligned with text during training and (2) capable of flexible attention to prevent the omission of relevant visual features.

\begin{figure*}[tbh]
  \centering
  \includegraphics[width=0.95\linewidth]{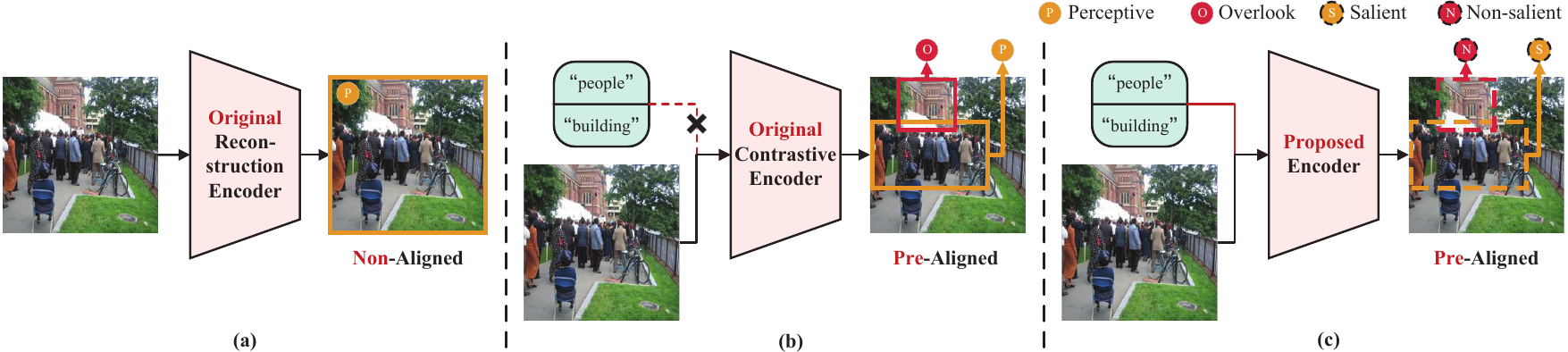}
  \vspace{-4mm} 
  \caption{\textbf{Motivation overview}. (a) The original reconstruction-based encoder perceives the full image but is \textit{not aligned with textual semantics}, thereby limiting its utility for LLMs in effectively interpreting the image embeddings. (b) The contrastive learning-based encoder only processes images \textit{without the benefit of textual instructions}, leading to a focus \textit{solely on salient objects} (\textcircled{P}) and neglecting user-specific concerns (\textcircled{O}). (c) Our proposed visual encoder addresses these limitations by flexibly adjusting its focus to highlight various objects, whether \textit{salient} (\textcircled{S}) or \textit{non-salient} (\textcircled{N}), according to the provided instructions.
  }
  \label{fig:atts_focus}
  \vspace{-5mm}
\end{figure*}

To address these challenges, we propose this \textbf{G}u\textbf{i}ding \textbf{V}isual \textbf{E}ncoder to perceive overlooked information (\textbf{GiVE}) approach, which aims to guide the visual encoder in adaptively adjusting its attention to well capture overlooked information. In this approach, we introduce a novel \textbf{Attention-Guided Adapter (AG-Adapter)} module that enhances the representation ability of the visual encoder by \textit{aligning the visual representations with abstract semantics}. This module also functions as a plug-in for generalizing abstract semantics, enabling it to more effectively address user queries. 
To tackle the above limitations in detail, GiVE incorporates another innovative module: \textbf{Object-focused Visual Semantic Learning}. This module employs three distinct model loss terms: (i) an Object-focused Image-Text Contrast (OITC) loss, which encourages the model to generate distinct embeddings for varied instructions, ensuring consideration of \textit{both salient and non-salient objects}; (ii) an Object-focused Image-Image Contrast (OIIC) loss, which improves the \textit{accuracy of object retrieval} by enabling the model to learn common features among in-class objects, thus enhancing concept generalization ability; and (iii) an Object-focused Image Discrimination (OID) loss, which improves the \textit{comprehensiveness of object retrieval} by facilitating the model in identifying specific objects and recognizing potential correlations among objects, thereby preventing the omission of objects. 
As illustrated in Fig. \ref{fig:atts_focus}, GiVE enables visual encoders to generate image embeddings that are rich in targeted information.

Our contributions can be summarized as follows:
\begin{itemize}
    \item We propose an innovative GiVE approach to give attention to overlooked information, enabling dynamic adjustment of focus by modifying textual instructions. This capability allows for flexible and adaptive attention to various objects within the image.
    \item We propose three novel loss functions: OITC, OIIC, and OID. The OITC loss addresses the current limitations of the original visual encoder, particularly its neglect of non-salient objects. The OIIC loss enhances ungeneralized representation ability, thereby improving object retrieval accuracy. The OID loss recognizes potential correlations among objects, further increasing the comprehensiveness of object retrieval. 
    \item We present a fine-grained image-text dataset with instructional labels, named the Multi-Object Instruction (MOInst) dataset, designed to provide semantic indications for different objects. Extensive experiments conducted on both constructed and real-world datasets showcase the effectiveness of our approach, achieving state-of-the-art performance.
\end{itemize}

\section{GiVE}

\begin{figure*}[tbh]
  \centering
  \includegraphics[width=0.8\linewidth]{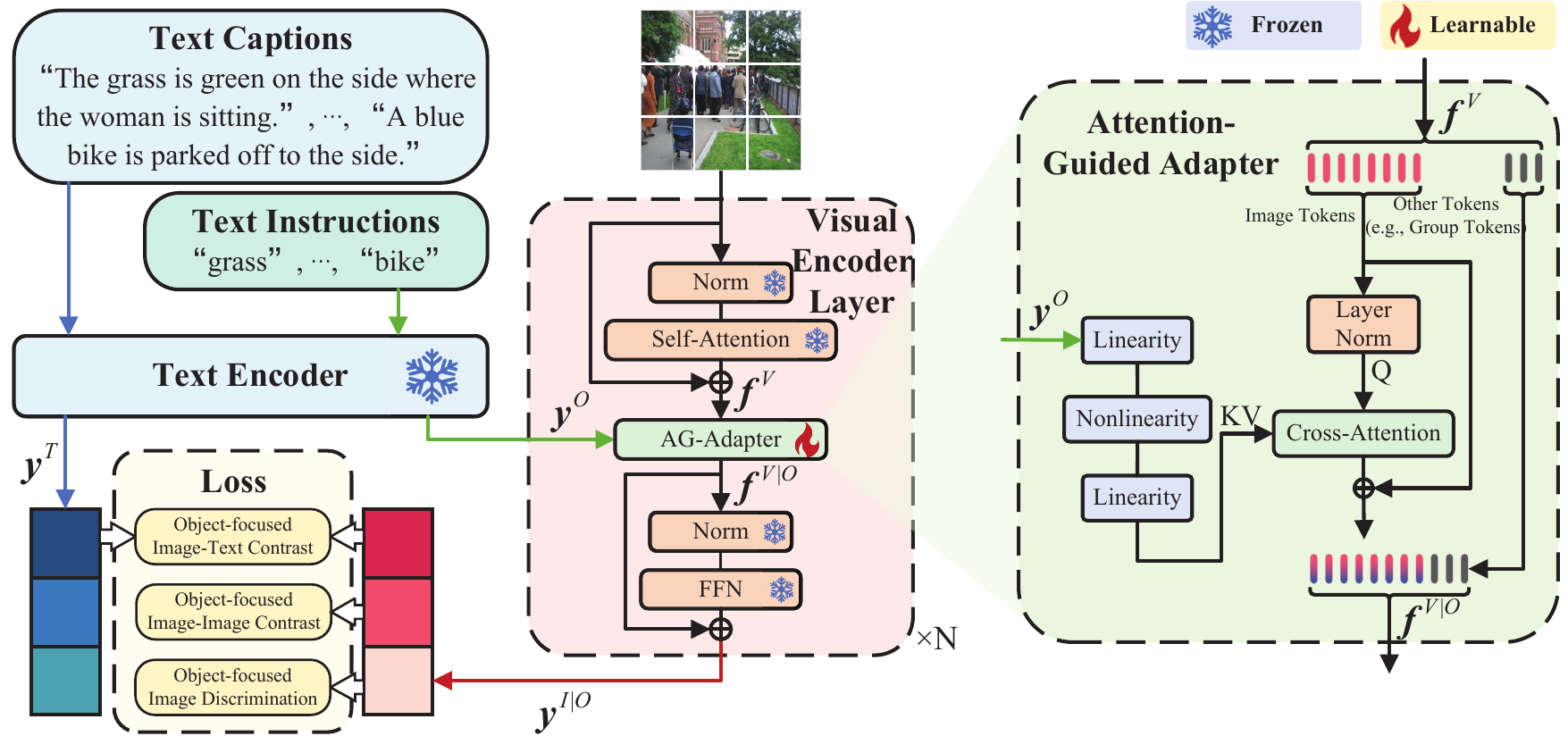}
  \vspace{-3mm}
  \caption{\textbf{Overall architecture of GiVE}. The plug-in module, AG-Adapter, is inserted into the feature extraction layers of the visual encoder and trained with the three losses proposed in our work: Object-focused Image-Text Contrast (OITC), Object-focused Image-Image Contrast (OIIC), and Object-focused Image Discrimination (OID). Cross-attention is used to emphasize the visual elements most relevant to the textual instructions. The text instructions are pre-integrated into the prompt template designated as ``a photo of \{object\}''.} \label{fig:architecture} 
  \vspace{-4mm}
\end{figure*}

\vspace{-1mm}
\subsection{Model Architecture}
\label{sec:arch}
\vspace{-1mm}
The overall architecture is depicted in Fig. \ref{fig:architecture}. 
The proposed model includes a visual encoder $\Phi_I(\cdot,\cdot)$ and a text encoder $\Phi_T(\cdot)$ to respectively encode visual and textual content. It accepts image-text-object triplets $\{(\boldsymbol{x}^{I}, \boldsymbol{x}^{T}, \boldsymbol{x}^{O})\}$ as input, where $\boldsymbol{x}^{I} \in \mathcal{I}$ is an image, $\boldsymbol{x}^{T} \in \mathcal{T}$ is a text, and $\boldsymbol{x}^{O} \in \mathcal{O}$ is an indicative text denoting the target object, such as ``person''. The model then extracts conditional image and text features $(\boldsymbol{y}^{I|O},\boldsymbol{y}^T)$ using paired encoders. When extracting conditional image features, the instruction feature $\boldsymbol{y}^O$ is fused with the visual data stream within the AG-Adapter module. Formally,
\vspace{-1mm}
\begin{equation}
    \boldsymbol{y}^T = \Phi_T(\boldsymbol{x}^{T}), \  \boldsymbol{y}^O = \Phi_T(\boldsymbol{x}^{O}), \ \boldsymbol{y}^{I|O} = \Phi_I(\boldsymbol{x}^{I}, \boldsymbol{y}^O),
    \vspace{-1mm}
\end{equation}
where $\boldsymbol{y}^T \in \mathbb{R}^d$ is text feature, $\boldsymbol{y}^O \in \mathbb{R}^d$ is instruction feature, and $\boldsymbol{y}^{I|O}$ is conditional image feature, i.e., the output of the visual encoder integrated with the AG-Adapter.
The AG-Adapter is trained using our designed Object-focused Visual Semantic Learning component containing three loss terms: OITC, OIIC, and OID. Note that during the training phase, the loss is calculated based on the output embeddings of both encoders. 
During the inference phase, the text encoder is retained to serialize the textual instructions. 

\begin{figure*}[tbh]
  \centering \includegraphics[width=0.95\linewidth]{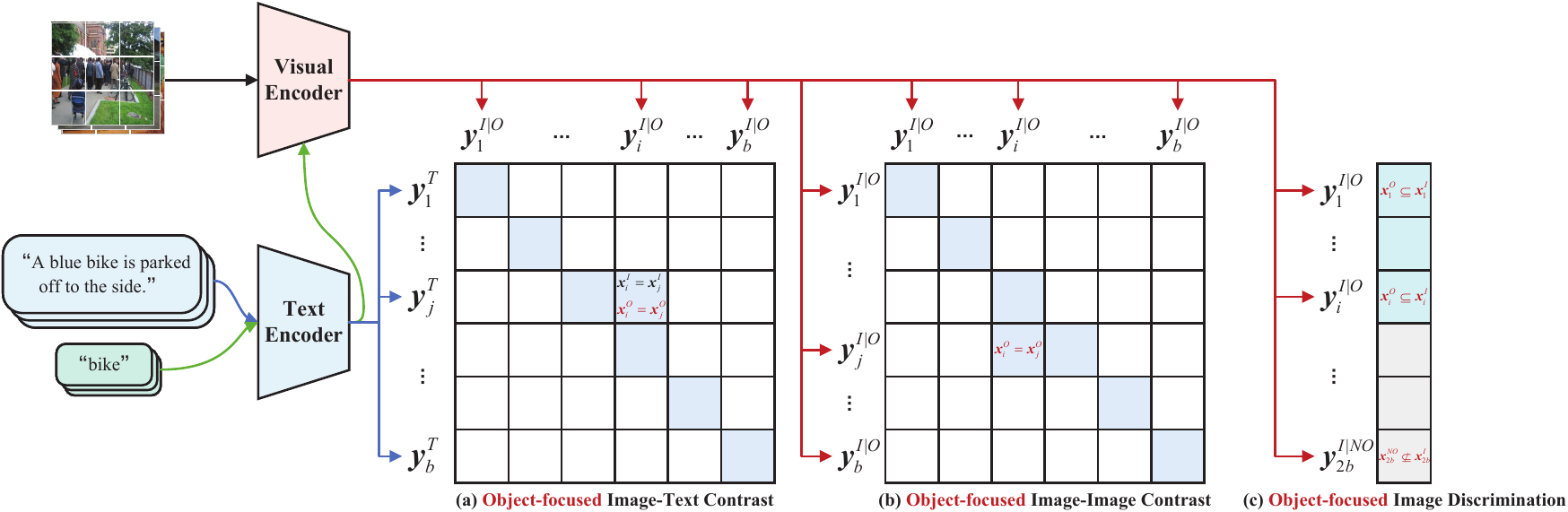}
  \vspace{-3mm}
  \caption{\textbf{Learning objectives illustration}. The image and text encoders jointly compute three losses. (a) For \textbf{Object-focused Image-Text Contrast}, the paired text and image should not only correspond to each other but also correspond to the same semantic object. (b) \textbf{Object-focused Image-Image Contrast} requires the model to predict pairs of image features that contain the same semantic object. (c) \textbf{Object-focused Image Discrimination} determines whether a specific object exists in the image or not. The text instructions, such as ``bike'', are pre-integrated into the prompt template designated as ``a photo of \{object\}''.}
  \label{fig:loss} 
  \vspace{-5mm}
\end{figure*}

\vspace{-1mm}
\subsection{Attention-Guided Adapter}
\label{sec:adapter}
\vspace{-1mm}
The proposed AG-Adapter module, as highlighted in the green rectangle in Fig. \ref{fig:architecture}, is a simple yet effective plug-in that interweaves semantic directives with visual cues, enabling the visual model to perceive queried objects. The AG-Adapter is inspired by the Latent Diffusion Model (LDM) \cite{ldm}, which enhances the alignment of visual representations with abstract semantics by conditioning on text representations. In particular, the AG-Adapter is incorporated into the pre-training feature extraction layers of the $\Phi_I(\cdot,\cdot)$, with only the inserted layer undergoing the training process.

Formally, in each layer of the visual encoder, the AG-Adapter module $\varphi(\cdot)$ enhances fine-grained object features:
\vspace{-3mm}
\begin{equation}     \boldsymbol{f}^{V|O}=\varphi(\boldsymbol{y}^O,\boldsymbol{f}^V),
\vspace{-2mm}
\end{equation}
where $\boldsymbol{y}^O$ is instruction feature derived from user queries and $\boldsymbol{f}^V$ refers to the feature of the visual data flow. The $\boldsymbol{f}^V \in \mathbb{R}^{(M+K) \times d}$ is composed of two types of tokens: image tokens $\boldsymbol{f}^I \in \mathbb{R}^{M \times d}$ and other tokens $\boldsymbol{f}^H \in \mathbb{R}^{K \times d}$. The image tokens are derived directly from the input image, while the role of other tokens depends on the baseline model. In the case of GroupViT \cite{groupvit}, other tokens are group tokens; in contrast, in CLIP \cite{clip}, they are absent, and $K=0$. Within the AG-Adapter, the dual-layer MLP serves to bridge the gap between text and image representations, namely,
\vspace{-2mm}
\begin{equation}
    \boldsymbol{f}^O = \text{MLP}(\boldsymbol{y}^O),
    \vspace{-3mm}
\end{equation}
where $\text{MLP}(\cdot)$ denotes two linear mappings and one non-linear mapping.
The image tokens are standardized in order to enhance training stability:
\vspace{-2mm}
\begin{equation}
    \hat{\boldsymbol{f}}^I = \text{Norm}(\boldsymbol{f}^I),
    \vspace{-3mm}
\end{equation}
where $\text{Norm}(\cdot)$ denotes the layer normalization operator. Subsequently, $\boldsymbol{f}^O$ and $\hat{\boldsymbol{f}}^I$ are merged through the cross-attention mechanism, with the resulting fused features integrated back into the visual data stream through residual concatenation as
\vspace{-2mm}
\begin{equation}   \boldsymbol{f}^{I|O}=\boldsymbol{f}^I+\text{Cross}(\hat{\boldsymbol{f}}^I,\boldsymbol{f}^O),
\vspace{-3mm}
\end{equation}
where $\text{Cross}(\cdot,\cdot)$ represents a cross-attention operator, $\hat{\boldsymbol{f}}^I$ is treated as query and $\boldsymbol{f}^O$ is key and value. $\boldsymbol{f}^{I|O}$ emphasizes the semantic objects while maintaining the integrity of the original visual information. The resulting tokens are then concatenated with other tokens to form the final features:
\vspace{-2mm}
\begin{equation}
    \boldsymbol{f}^{V|O} = [\boldsymbol{f}^{I|O} ; \boldsymbol{f}^H],
    \vspace{-3mm}
\end{equation}
where $[\cdot;\cdot]$ represents the concatenation of two vectors.
The feature $\boldsymbol{f}^{V|O}$ that enriches the visual information of the indicated object is fed to the subsequent module.

In this module, we counter-intuitively use the image as the query and the text as the key and value. This approach is driven by two considerations. First, it ensures that visual information is adequately preserved. Typically, instruction tokens are much shorter than image tokens, so a text-based query might not carry sufficient information. Second, the LDM \cite{ldm} conditions the image feature delivery on the text, also using the image as the query and the text as the key and value, indicating that such modal fusion is feasible. Our subsequent experiments confirm the effectiveness of this role assignment.

\vspace{-1mm}
\subsection{Object-Focused Visual Semantic Learning}
\label{sec:loss}
\vspace{-1mm}
During the abstract concept learning phase, the original parameters of the image and text encoders are frozen, and the parameters of the AG-Adapter are trained. 
In each training batch, we sample $b$ image-text-object triples $\{(\boldsymbol{x}^{I}_k, \boldsymbol{x}^{T}_k, \boldsymbol{x}^{O}_k)\}^b_{k=1}$ and encode them to obtain image-text feature pairs $\{(\boldsymbol{y}^{I|O}_k, \boldsymbol{y}^{T}_k)\}^b_{k=1}$. 
Our goal is to train the AG-Adapter to extract the most relevant visual representations based on semantic instructions. As shown in Fig. \ref{fig:loss}, to achieve this goal, we jointly optimize three training objectives that share model parameters.

\textbf{Object-focused Image-Text Contrast (OITC)} objective is designed to align image and text representations centered around instructed objects, with the aim of maximizing their mutual information. This objective requires the visual encoder to generate distinct features for different instructions.
The similarity between features is computed as follows:
\vspace{-2mm}
\begin{equation}
    s^{I|O}_{i,j} = s^{T}_{i,j} = (\boldsymbol{y}^{I|O}_i)^\top \boldsymbol{y}^{T}_j,
    \vspace{-3mm}
\end{equation}
where $s^{I|O}_{i,j}$ denotes the similarity of the $i$-th conditional image feature to the $j$-th text feature, and $s^{T}_{i,j}$ represents the similarity between the $i$-th text feature and the $j$-th text feature.
For the conditional image feature $\boldsymbol{y}^{I|O}_k$ in feature pair $\{(\boldsymbol{y}^{I|O}_k, \boldsymbol{y}^{T}_k)\}$, the corresponding text features $\boldsymbol{y}^{T}_{j|\boldsymbol{x}^{I}_j=\boldsymbol{x}^{I}_k \land \boldsymbol{x}^{O}_j=\boldsymbol{x}^{O}_k}$ of the same object within the same image are positive, while other text features within the batch are negative. The loss of a batch can be represented by
\begin{align}    &\mathcal{L}^{I|O}_k(\boldsymbol{y}^{I|O}_k,\{\boldsymbol{y}^T_j\}^b_{j=1})= \notag \\
    &-\frac{1}{b} \log\frac{\sum_{k,j} \exp(s^{I|O}_{k,j} \mid \boldsymbol{x}^{I}_j=\boldsymbol{x}^{I}_k \land \boldsymbol{x}^{O}_j=\boldsymbol{x}^{O}_k)}{\sum_j \exp(s^{I|O}_{k,j})}, 
\end{align}
\vspace{-1mm}
\begin{align}    
    &\mathcal{L}^T_k(\boldsymbol{y}^{T}_k,\{\boldsymbol{y}^{I|O}_j\}^b_{j=1})= \notag \\
    &-\frac{1}{b} \log \frac{\sum_{k,j} \exp(s^{T}_{k,j} \mid \boldsymbol{x}^{I}_j=\boldsymbol{x}^{I}_k \land \boldsymbol{x}^{O}_j=\boldsymbol{x}^{O}_k)}{\sum_j \exp(s^{T}_{k,j})}, 
    \vspace{-1mm}
\end{align} 
\vspace{-1mm}
\begin{equation}
    \mathcal{L}^\text{OITC}=\frac{1}{2} \sum^b_{k=1}(\mathcal{L}^{I|O}_k + \mathcal{L}^T_k),
    \vspace{-1mm}
\end{equation}
where $\mathcal{L}^{I|O}_k$ is image-to-text contrastive loss, $\mathcal{L}^T_k$ is text-to-image contrastive loss, and $\mathcal{L}^\text{OITC}$ is total loss.

\textbf{Object-focused Image-Image Contrast (OIIC)} loss emphasizes the commonality of objects within the same class, requiring the encoder to generate similar features for these objects. The contrast is performed within the image. For a feature $\boldsymbol{y}^{I|O}_k$, $\boldsymbol{y}^{I|O}_{j|\boldsymbol{x}^{O}_j=\boldsymbol{x}^{O}_k}$ is its positive example, while other conditional image features in the batch serve as in-batch negatives. The similarity computation of conditional image features is expressed as
\vspace{-1mm}
\begin{equation}
    s_{i,j} = (\boldsymbol{y}^{I|O}_i)^\top \boldsymbol{y}^{I|O}_j.
    \vspace{-1mm}
\end{equation}
The OIIC loss, denoted by $\mathcal{L}^\text{OIIC}$, can be represented as
\vspace{-1mm}
\begin{equation}    
    \mathcal{L}^\text{OIIC} = -\frac{1}{b} \sum^b_{k=1}\mathcal \log \frac{\sum_{k,j} \exp(s_{k,j} \mid \boldsymbol{x}^{O}_j=\boldsymbol{x}^{O}_k)}{\sum_j \exp(s_{k,j})}.
    \vspace{-1mm}
\end{equation}

\renewcommand{\arraystretch}{0.8}
\begin{table}[tb]
  \centering
  \caption{Top-1 F1 and AUC (\%) of zero-shot image classification on multi-object datasets}
  \vspace{-2mm}
  \label{tab:classification}
  \begingroup
    \begin{threeparttable}
    \setlength{\tabcolsep}{9pt}
    \begin{tabular}{>{\small}l>{\small}c>{\small}c>{\small}c>{\small}c}
    \toprule
    \multirow{2}{*}{Model}   & \multicolumn{2}{c}{LVIS$^{f}$}     & \multicolumn{2}{c}{LVIS$^{a}$} \\
             & F1 & AUC    & F1 & AUC \\
    \midrule
    Instruct.\footnotemark[1]  & -0.1\footnotemark[4] & 49.9 & 0.1 & 50.0 \\
    \midrule
    \rowcolor{gray!20}
    CLIP-ViT    & 11.3  & 57.8   & 8.6   & 55.5 \\
    + GiVE & 50.7 & 75.6 & 56.2 & 77.3 \\
    Improv. & \textcolor{ForestGreen}{348.7\%} & \textcolor{ForestGreen}{30.8\%} & \textcolor{ForestGreen}{553.5\%} & \textcolor{ForestGreen}{39.3\%} \\
    \midrule
    \rowcolor{gray!20}
    GroupViT$_y$\footnotemark[2]  & 9.6   & 56.8 & 7.3   & 54.8 \\
    + GiVE & 31.9 & 65.3 & 42.3 & 69.8 \\
    Improv. & \textcolor{ForestGreen}{232.3\%} & \textcolor{ForestGreen}{15.0\%} & \textcolor{ForestGreen}{479.5\%} & \textcolor{ForestGreen}{27.4\%} \\
    \midrule
    \rowcolor{gray!20}
    GroupViT$_r$\footnotemark[3]  & 10.1  & 57.2  & 8.2   & 55.4 \\
    + GiVE & 31.8 & 65.9 & 42.8 & 70.3 \\
    Improv. & \textcolor{ForestGreen}{214.9\%} & \textcolor{ForestGreen}{15.2\%} & \textcolor{ForestGreen}{422.0\%} & \textcolor{ForestGreen}{26.9\%} \\
    \midrule
    \rowcolor{gray!20}
    SigLIP  & 10.7  & 57.3 & 9.0  & 55.6 \\
    + GiVE & 48.6 & 74.2 & 53.0 & 75.0 \\
    Improv. & \textcolor{ForestGreen}{354.2\%} & \textcolor{ForestGreen}{29.5\%} & \textcolor{ForestGreen}{488.9\%} & \textcolor{ForestGreen}{34.9\%} \\
    \midrule
    \rowcolor{gray!20}
    OwlViT & 0.0  & 50.0 & 0.0  & 50.0 \\
    + GiVE & 40.6 & 70.6 & 41.8 & 69.9 \\
    Improv. & \textcolor{ForestGreen}{--} & \textcolor{ForestGreen}{41.2\%} & \textcolor{ForestGreen}{--} & \textcolor{ForestGreen}{39.8\%} \\
    \bottomrule
    \end{tabular}%
    \begin{tablenotes}
        \item[1] Classification is based on the instruction text rather than the image.
        \item[2,3] GroupViT-gcc-yfcc and GroupViT-gcc-redcaps, denote two variants of GroupViT trained with different datasets, respectively.
        \item[4] Since our evaluation metric accounts for and subtracts the influence of textual interference, it can result in negative values.
    \end{tablenotes}
    \end{threeparttable}
    \endgroup
    \vspace{-5mm}
\end{table}%

\begin{table}[tbh]
  \centering
  \caption{Fine-tuned image-text retrieval results on MOInst dataset}
  \vspace{-2mm}
  \label{tab:retrieval}
  \begingroup
    \setlength{\tabcolsep}{4pt}
    \begin{threeparttable}
    \begin{tabular}{llcccc}
    \toprule
    \multirow{2}{*}{Model} & \multirow{2}{*}{\makecell[l]{\#Param.\footnotemark[1]}}& \multicolumn{2}{c}{Image $\rightarrow$ Text} & \multicolumn{2}{c}{Text $\rightarrow$ Image} \\
          &       & R@1 & R@5 & R@1 & R@5 \\
    \midrule
    Instruct. & & 0.1 & 0.1 & -0.2 & 0.4 \\
    \midrule
    \rowcolor{gray!20}
    CLIP-ViT & 88M & 7.6 & 18.7 & 5.4 & 19.7 \\
    + GiVE & 62M & 29.1 & 53.3 & 29.8 & 54.9 \\
    Improv. &  & \textcolor{ForestGreen}{282.9\%} & \textcolor{ForestGreen}{185.0\%} & \textcolor{ForestGreen}{451.9\%} & \textcolor{ForestGreen}{178.7\%} \\
    \midrule
    \rowcolor{gray!20}
    GroupViT$_{y}$ & 31M   & 5.9 & 18.0 & 4.2 & 17.8 \\
    + GiVE & 15M & 13.2 & 29.3 & 16.7 & 34.9 \\
    Improv. &  & \textcolor{ForestGreen}{123.7\%} & \textcolor{ForestGreen}{62.8\%} & \textcolor{ForestGreen}{297.6\%} & \textcolor{ForestGreen}{96.1\%} \\
    \midrule
    \rowcolor{gray!20}
    GroupViT$_{r}$ & 31M   & 7.2 & 19.1 & 4.7 & 18.7 \\
    + GiVE & 15M & 12.6 & 29.8 & 17.1 & 36.2 \\
    Improv. &  & \textcolor{ForestGreen}{75.0\%} & \textcolor{ForestGreen}{56.0\%} & \textcolor{ForestGreen}{263.8\%} & \textcolor{ForestGreen}{93.6\%} \\
    \midrule
    \rowcolor{gray!20}
    SigLIP & 93M  & 9.3 & 23.6 & 6.9 & 25.6 \\
    + GiVE & 85M & 20.7 & 43.6 & 25.7 & 48.1 \\
    Improv. &  & \textcolor{ForestGreen}{122.6\%} & \textcolor{ForestGreen}{84.7\%} & \textcolor{ForestGreen}{272.5\%} & \textcolor{ForestGreen}{87.9\%} \\
    \midrule
    \rowcolor{gray!20}
    OwlViT & 88M  & 5.6 & 16.5 & 4.1 & 16.5 \\
    + GiVE & 62M & 12.4 & 32.1 & 14.4 & 35.2 \\
    Improv. &  & \textcolor{ForestGreen}{121.4\%} & \textcolor{ForestGreen}{94.5\%} & \textcolor{ForestGreen}{251.2\%} & \textcolor{ForestGreen}{113.3\%} \\
    \bottomrule
    \end{tabular}%
    \begin{tablenotes}
        \item[1] The number of trainable parameters.
    \end{tablenotes}
    \end{threeparttable}
    \endgroup
    \vspace{-5mm}
\end{table}%

\textbf{Object-focused Image Discrimination (OID)} is a binary classification task that requires the model to predict whether a given image and the indicated object match. For each batch of sample pairs $\{(\boldsymbol{x}^{I}_k, \boldsymbol{x}^{O}_k)\}^b_{k=1}$, we additionally construct $b$ negative pairs $\{(\boldsymbol{x}^{I}_k, \boldsymbol{x}^{NO}_k)\}^b_{k=1}$, where $\boldsymbol{x}^{NO}_k$ indicating object not appear in $\boldsymbol{x}^{I}_k$. These positive and negative samples $\{(\boldsymbol{x}^{I}_k, \boldsymbol{x}^{O \cup NO}_k)\}^{2b}_{k=1}$ are encoded to $\{\boldsymbol{y}^{I|O \cup NO}_k\}^{2b}_{k=1}$, which are then input into a binary linear classifier to obtain logits $\{\boldsymbol{z}_k\}^{2b}_{k=1}$. The loss function is formalized as
\vspace{-1mm}
\begin{align}
    p_i &= \frac{1}{1+\exp(-z_i)}, \\
    \mathcal{L}^\text{OID} &= -\frac{1}{2b} \sum_{i=1}^{2b} [ t_i \log(p_i) + (1-t_i) \log(1-p_i) ],
    \vspace{-1mm}
\end{align}
where $p_i \in \mathbb{R}$ and $t_i \in \{0, 1\}$ denote the predicted probability and true label of the $i$-th sample, respectively.

\vspace{-2mm}
\subsection{Instruction of New Dataset}
\vspace{-1mm}
Although the Visual Genome (VG) dataset \cite{vg} provides multiple objects and captions per image, it lacks object-caption correspondence, features noisy labels, and contains low-quality text. Therefore, it is not recommended to use the VG dataset directly. To address these issues, we construct the Multi-Object Instruction (MOInst) dataset.
This dataset comprises 81,536 high-fidelity, complex images, accompanied by 244,378 textual captions. Each caption is associated with one of 264 distinct categories.
\begin{table*}[tbh]
  \centering
  \caption{The effect of semantic instructions on secondary coding methods. AG-CLIP is CLIP-ViT with GiVE for attention guidance}
  \vspace{-2mm}
  \label{tab:post-processing}
  \begingroup
    \setlength{\tabcolsep}{4.9pt}
    \begin{threeparttable}        
    \begin{tabular}{>{\small}l>{\small}l>{\small}l>{\small}l>{\small}l>{\small}l>{\small}c>{\small}c>{\small}c>{\small}c>{\small}c>{\small}c>{\small}c}
    \toprule
    \multirow{3}{*}{Model} & \multirow{3}{*}{\#Param.} & \multirow{3}{*}{Inst.} & \multirow{3}{*}{Img Enc.} & \multirow{3}{*}{Type} & \multirow{3}{*}{GiVE} & \multirow{3}{*}{\makecell[l]{Post Proc.}} & \multicolumn{2}{c}{MOInst}  & \multicolumn{2}{c}{LVIS$^f$} & \multicolumn{2}{c}{LVIS$^a$} \\
          &    & & & &    &   & I2T   & T2I  & \multirow{2}{*}{F1} & \multirow{2}{*}{AUC} & \multirow{2}{*}{F1} & \multirow{2}{*}{AUC} \\
          &    & & &  &   &    & R@1  & R@1 &  &  &  &  \\
    \midrule
    \rowcolor{gray!20}
    BLIP-2  & 1.2B & \XSolidBrush & EVA-CLIP-ViT & G/14 & \XSolidBrush & Q-Former & 14.1 & 9.4 & 14.1 & 58.2 & 11.3 & 56.3 \\
    BLIP-2  & 388M & \CheckmarkBold & CLIP-ViT & B/32& \CheckmarkBold & Q-Former & \textbf{30.1}  & \textbf{31.1}  & 35.2  & 67.4  & 40.1  & 68.9 \\
    Improv. & &&&& &  & \textcolor{ForestGreen}{113.5\%} & \textcolor{ForestGreen}{230.9\%} & \textcolor{ForestGreen}{149.6\%} & \textcolor{ForestGreen}{15.8\%} & \textcolor{ForestGreen}{254.9\%} & \textcolor{ForestGreen}{22.4\%} \\
    \midrule
    \rowcolor{gray!20}
    InstructBLIP  & 1.2B & \CheckmarkBold & EVA-CLIP-ViT & G/14 & \XSolidBrush & Q-Former & 19.8 & 22.1 & 28.6 & 64.7 & 23.0 & 60.9 \\
    InstructBLIP & \makecell[l]{388M} & \makecell[l]{\CheckmarkBold} & \makecell[l]{CLIP-ViT} & \makecell[l]{B/32} & \makecell[l]{\CheckmarkBold} & \makecell[l]{Q-Former}  & 24.2  & 21.9  & 35.0  &  67.1 & 42.7  & 70.6  \\
    Improv. &  &&&&&  & \textcolor{ForestGreen}{22.2\%} & \textcolor{BrickRed}{-0.9\%} & \textcolor{ForestGreen}{22.4\%} & \textcolor{ForestGreen}{3.7\%} & \textcolor{ForestGreen}{85.7\%} & \textcolor{ForestGreen}{15.9\%} \\
    \midrule
    AG-CLIP & 62M & \CheckmarkBold & CLIP-ViT & B/32 & \CheckmarkBold & -- & 29.1 & 29.8 & \textbf{50.7} & \textbf{75.6} & \textbf{56.2} & \textbf{77.3} \\
    \bottomrule
    \end{tabular}%
    \end{threeparttable}
    \endgroup  
    \vspace{-6mm}
\end{table*}%

\section{Experiments}
\subsection{Experimental Setup}
\subsubsection{Datasets}
We evaluate the effectiveness of GiVE using both the LVIS dataset \cite{lvis} and MOInst dataset, each annotating multiple objects per image. The comprehensive LVIS dataset, referred to as LVIS$^a$, comprises 1,203 categories, from which, we derive a subset, denoted as LVIS$^f$, with 405 ``frequent'' categories. In contrast, the MOInst dataset used for training contains 264 categories, with category label overlaps of 7.7\% and 17.6\% with the two LVIS datasets, respectively.
We also conduct experiments on the COCO \cite{coco}, SUN397 \cite{sun397}, and ImageNet \cite{imagenet} datasets.

\subsubsection{Baselines}
We evaluate the gains brought by integrating the GiVE with several representative ViT baselines, including CLIP \cite{clip}, GroupViT \cite{groupvit}, SigLIP \cite{siglip}, OwlViT \cite{owlvit}, and MetaCLIP \cite{metaclip}. We also apply the GiVE to larger encoding frameworks, such as BLIP-2 \cite{blip-2} and InstructBLIP \cite{instructblip}. Throughout this paper, unless explicitly stated otherwise, the experiments with GiVE are conducted on the CLIP platform, with the unmarked CLIP model referring to CLIP-ViT-B/32.

\subsubsection{Evaluation Metrics}
Following previous research \cite{filip}, we evaluate image classification and image-text retrieval tasks using scores based solely on vector similarity, avoiding further optimization to accurately reflect the extracted feature information. To focus on the capabilities of visual encoders, we quantify and mitigate the influence of textual data in methods like InstructBLIP \cite{instructblip}.

\begin{table}[tb]
  \centering
    \caption{Ablation study}
    \vspace{-1mm}
    \label{tab:ablation}
    \begingroup
    \setlength{\tabcolsep}{3pt}
    \begin{threeparttable}
    \begin{tabular}{>{\small}l>{\small}l>{\small}l>{\small}c>{\small}c>{\small}c>{\small}c>{\small}c>{\small}c>{\small}c}
    \toprule
    \multirow{3}{*}{Loss} & \multirow{3}{*}{Fusion\footnotemark[1]} & \multirow{3}{*}{Inst.} & \multicolumn{2}{c}{MOInst} & \multicolumn{2}{c}{LVIS$^f$} & \multicolumn{2}{c}{LVIS$^a$} \\
          &       &       & I2T   & T2I   & \multirow{2}{*}{F1} & \multirow{2}{*}{AUC} & \multirow{2}{*}{F1} & \multirow{2}{*}{AUC} \\
          &       &       & R@1  &  R@1 &   &    &   &    &  \\
    \midrule
    \rowcolor{gray!20}
    \XSolidBrush & \XSolidBrush & \XSolidBrush & 7.6 & 5.4 & 11.3 & 57.8 & 8.6 & 55.5 \\
    \midrule
    -OITC\footnotemark[2] & \makecell[l]{Early\footnotemark[3]+late\footnotemark[4]}   & \CheckmarkBold     &  0.1 & 0.1 & 0.3 & 50.0 & 0.1 & 50.0 \\
    -OIIC & \makecell[l]{Early+late \\ }   & \CheckmarkBold     & 28.1 & 29.4 & 48.4 & 74.0 & 55.7 & 76.7 \\
    -OID & \makecell[l]{Early+late \\ }   & \CheckmarkBold     & 28.1 & 29.7 & 37.6 & 67.8 & 47.6 & 72.2 \\
    \midrule
    All   & Early & \CheckmarkBold     &  6.4 & 8.2 & 27.7 & 63.6 & 37.2 & 67.3 \\
    All   & Late  & \CheckmarkBold     & 27.1 & 29.1 & 47.4 & 73.6 & 55.6 & 76.5 \\
    All   & Sparse\footnotemark[5] & \CheckmarkBold     & 27.0 & 29.0 & 46.9 & 73.0 & 55.1 & 76.3 \\
    \midrule
    All   & \makecell[l]{Early+late \\} & \XSolidBrush     & 0.0 & 0.1 & 0.3 & 50.0 & 0.4 & 50.0 \\
    \midrule
    All   & \makecell[l]{Early+late \\ (dense\footnotemark[6])}   & \CheckmarkBold     & \textbf{29.1} & \textbf{29.8} & \textbf{50.7} & \textbf{75.6} & \textbf{56.2} & \textbf{77.3} \\
    \bottomrule
    \end{tabular}%
    \begin{tablenotes}
        \item[1] The layer where the AG-Adapter is inserted.
        \item[2] ``-'' indicates the elimination of the loss function during training.
        \item[3,4,5,6] ``Early'', ``late'', ``sparse'', and ``dense'' indicate that features are fused in the first half of layers, the latter half of layers, alternate layers, and all layers, respectively.
    \end{tablenotes}
    \end{threeparttable}
    \endgroup
    \vspace{-2mm}
\end{table}%

\subsection{Performance Evaluation}
\subsubsection{Image Classification}
We conduct zero-shot image classification on the ``frequent'' subset and the full LVIS test set. Table \ref{tab:classification} presents the results on five ViT baselines.  Key observations from these experiments are as follows:

\begin{itemize}
    \item The evaluation value for the instruction text is nearly equivalent to random categorization, demonstrating that our metric successfully filters out textual interference. This outcome ensures that our work fairly compares the visual feature extraction capabilities of each model.
    \item The GiVE demonstrates a notable improvement in all baselines across all metrics on both evaluation datasets. This highlights the efficacy of the GiVE's capacity to redirect attention based on semantic instructions.
    \item OwlViT, a ViT designed for object detection, requires fine-tuning to serve effectively as an image feature extractor. However, GiVE can unlock its potential.
    \item The AG-Adapter, trained on the MOInst dataset with 264 classes, achieves F1 scores of over 40\% on the LVIS$^a$ dataset with 1,203 classes. The observed gaps in category magnitudes indicate that the instruction semantics have generalizability beyond the training scope. 
\end{itemize}

\subsubsection{Image-Text Retrieval}
Image-text retrieval includes two subtasks: image-to-text retrieval and text-to-image retrieval. We evaluate the models on the MOInst dataset. Since the AG-Adapter is trained on this dataset, we compare the results under the fine-tuned setting, as recorded in Table \ref{tab:retrieval}. 
It is easy to see that GiVE outperforms all baseline methods while utilizing fewer trainable parameters. The informativeness gap between instruction and caption indicates that simple instruction alone is insufficient for the model to achieve such a high hit rate. Together, GiVE effectively extracts target visual features. Additionally, the evaluation results for ``Instruct.'' further support the credibility of our experimental findings.

\subsubsection{Comparison with Secondary Coding Methods}
The BLIP-2 and InstructBLIP methods utilize a Q-Former to re-encode image features after the visual encoder, with InstructBLIP incorporating text instructions into the Q-Former. We replace the visual encoders of these two methods with AG-CLIP-ViT-B/32 and then fine-tune them on MOInst following their respective strategies. The results are shown in Table \ref{tab:post-processing}, where ``Inst.'' indicates whether the model receives additional instructions. ``Img Enc.'' specifies the image encoder used by the model, and ``Type'' denotes the type of image encoder. ``GiVE'' shows whether our GiVE method is applied to the image encoder, and ``Post Proc.'' identifies the type of post-processor used for the secondary encoding of the image embeddings. From the table, it can be observed that:

\begin{itemize}
    \item Our AG-CLIP shows performance improvements compared to the original BLIP-2 and InstructBLIP, despite the significant difference in the number of parameters. This supports the validity of injecting instructions.
    \item Semantic instructions prove to be more efficient during visual feature extraction than when applied post-process. In our second set of experiments, we replace the original giant EVA-CLIP in InstructBLIP with the base AG-CLIP, resulting in superior performance across most metrics. Although AG-CLIP performs slightly worse in the text-to-image retrieval task, the substantial difference in the number of training parameters makes this acceptable.
    \item Secondary encoding may weaken the abstract semantics in the visual features, leading to decreased performance in image classification tasks, as evidenced by the superior performance of AG-CLIP compared to BLIP-2 and InstructBLIP on LVIS$^f$ and LVIS$^a$ datasets.
\end{itemize}

\subsection{Ablation Studies}

We perform an ablation study of GiVE from three aspects: loss, fusion layers, and abstract semantic instructions. For instructions, we replace short object prompts with detailed descriptions to remove abstract semantics. Table \ref{tab:ablation} shows the results, with the gray row for CLIP-ViT and the last row for AG-CLIP-ViT. The main observations are as follows:

\begin{itemize}
    \item The ablation study on loss functions suggests that all three losses are crucial, with particular emphasis on the OITC loss, which is responsible for aligning object-focused visual features with text features.
    \item The late encoding layers have a significantly greater impact on abstract semantics compared to the early and sparse layers, though the early and sparse layers also contribute to the understanding of semantics.
    \item In the absence of instructions, captions can serve as textual inputs. However, overly specific captions may cause the visual encoder to rely heavily on textual instructions,  possibly missing key details in the image.
\end{itemize}

\section{Conclusion}
This paper presents GiVE, a novel approach enhancing visual encoders' integration with LLMs by addressing semantic alignment and overlooked information. GiVE features the AG-Adapter and three innovative loss functions—OITC, OIIC, and OID. The AG-Adapter aligns visual representations with abstract semantics, while the OITC loss ensures attention to both salient and non-salient objects, and the OIIC and OID losses enhance object retrieval accuracy and comprehensiveness. Experiments show GiVE's significant improvements over existing methods in multiple tasks.

\bibliographystyle{IEEEbib}

\bibliography{icme2025references}

\begin{thebibliography}{10}

\bibitem{blip-2}
Junnan Li, Dongxu Li, Silvio Savarese, and Steven C.~H. Hoi,
\newblock ``{BLIP-2:} bootstrapping language-image pre-training with frozen image encoders and large language models,''
\newblock in {\em {ICML}}, 2023, pp. 19730--19742.

\bibitem{videoldm}
Andreas Blattmann, Robin Rombach, Huan Ling, Tim Dockhorn, Seung~Wook Kim, Sanja Fidler, and Karsten Kreis,
\newblock ``Align your latents: High-resolution video synthesis with latent diffusion models,''
\newblock in {\em {CVPR}}, 2023, pp. 22563--22575.

\bibitem{Amanpreet_VQA}
Amanpreet Singh, Vivek Natarajan, Meet Shah, Yu~Jiang, Xinlei Chen, Dhruv Batra, Devi Parikh, and Marcus Rohrbach,
\newblock ``Towards {VQA} models that can read,''
\newblock in {\em {CVPR}}, 2019, pp. 8317--8326.

\bibitem{zhou2025pa}
Sen Wang, Dongliang Zhou, Liang Xie, Chao Xu, Ye~Yan, and Erwei Yin,
\newblock ``Panogen++: Domain-adapted text-guided panoramic environment generation for vision-and-language navigation,''
\newblock {\em Neural Networks}, pp. 1--1, 2025.

\bibitem{llava-1.5}
Haotian Liu, Chunyuan Li, Yuheng Li, and Yong~Jae Lee,
\newblock ``Improved baselines with visual instruction tuning,''
\newblock {\em CoRR}, vol. abs/2310.03744, 2023.

\bibitem{vqvae}
A{\"{a}}ron van~den Oord, Oriol Vinyals, and Koray Kavukcuoglu,
\newblock ``Neural discrete representation learning,''
\newblock in {\em {NIPS}}, 2017, pp. 6306--6315.

\bibitem{mirl}
Guoxi Huang, Hongtao Fu, and Adrian~G. Bors,
\newblock ``Masked image residual learning for scaling deeper vision transformers,''
\newblock in {\em NeurIPS}, 2023.

\bibitem{survey_mllm}
Shukang Yin, Chaoyou Fu, Sirui Zhao, Ke~Li, Xing Sun, Tong Xu, and Enhong Chen,
\newblock ``A survey on multimodal large language models,''
\newblock {\em CoRR}, vol. abs/2306.13549, 2023.

\bibitem{zhou2023fcboost}
Dongliang Zhou, Haijun Zhang, Jianghong Ma, Jicong Fan, and Zhao Zhang,
\newblock ``Fcboost-net: A generative network for synthesizing multiple collocated outfits via fashion compatibility boosting,''
\newblock in {\em Proceedings of the 31st ACM international conference on multimedia}, 2023, pp. 7881--7889.

\bibitem{clip}
Alec Radford, Jong~Wook Kim, Chris Hallacy, Aditya Ramesh, Gabriel Goh, Sandhini Agarwal, Girish Sastry, Amanda Askell, Pamela Mishkin, Jack Clark, Gretchen Krueger, and Ilya Sutskever,
\newblock ``Learning transferable visual models from natural language supervision,''
\newblock in {\em {ICML}}, 2021, vol. 139, pp. 8748--8763.

\bibitem{zhou2023bc}
Dongliang Zhou, Haijun Zhang, Jianghong Ma, and Jianyang Shi,
\newblock ``Bc-gan: A generative adversarial network for synthesizing a batch of collocated clothing,''
\newblock {\em IEEE Transactions on Circuits and Systems for Video Technology}, vol. 34, no. 5, pp. 3245--3259, 2023.

\bibitem{ldm}
Robin Rombach, Andreas Blattmann, Dominik Lorenz, Patrick Esser, and Bj{\"{o}}rn Ommer,
\newblock ``High-resolution image synthesis with latent diffusion models,''
\newblock in {\em {CVPR}}, 2022, pp. 10674--10685.

\bibitem{groupvit}
Jiarui Xu, Shalini~De Mello, Sifei Liu, Wonmin Byeon, Thomas~M. Breuel, Jan Kautz, and Xiaolong Wang,
\newblock ``Groupvit: Semantic segmentation emerges from text supervision,''
\newblock in {\em {CVPR}}, 2022, pp. 18113--18123.

\bibitem{vg}
Ranjay Krishna, Yuke Zhu, Oliver Groth, Justin Johnson, Kenji Hata, Joshua Kravitz, Stephanie Chen, Yannis Kalantidis, Li{-}Jia Li, David~A. Shamma, Michael~S. Bernstein, and Li~Fei{-}Fei,
\newblock ``Visual genome: Connecting language and vision using crowdsourced dense image annotations,''
\newblock {\em Int. J. Comput. Vis.}, vol. 123, no. 1, pp. 32--73, 2017.

\bibitem{lvis}
Agrim Gupta, Piotr Doll{\'{a}}r, and Ross~B. Girshick,
\newblock ``{LVIS:} {A} dataset for large vocabulary instance segmentation,''
\newblock in {\em {CVPR}}, 2019, pp. 5356--5364.

\bibitem{coco}
Tsung{-}Yi Lin, Michael Maire, Serge~J. Belongie, James Hays, Pietro Perona, Deva Ramanan, Piotr Doll{\'{a}}r, and C.~Lawrence Zitnick,
\newblock ``Microsoft {COCO:} common objects in context,''
\newblock in {\em {ECCV}}, 2014, vol. 8693, pp. 740--755.

\bibitem{sun397}
Jianxiong Xiao, James Hays, Krista~A. Ehinger, Aude Oliva, and Antonio Torralba,
\newblock ``{SUN} database: Large-scale scene recognition from abbey to zoo,''
\newblock in {\em {CVPR}}, 2010, pp. 3485--3492.

\bibitem{imagenet}
Shanghua Gao, Zhong{-}Yu Li, Ming{-}Hsuan Yang, Ming{-}Ming Cheng, Junwei Han, and Philip H.~S. Torr,
\newblock ``Large-scale unsupervised semantic segmentation,''
\newblock {\em {IEEE} Trans. Pattern Anal. Mach. Intell.}, vol. 45, no. 6, pp. 7457--7476, 2023.

\bibitem{siglip}
Xiaohua Zhai, Basil Mustafa, Alexander Kolesnikov, and Lucas Beyer,
\newblock ``Sigmoid loss for language image pre-training,''
\newblock in {\em {ICCV}}, 2023, pp. 11941--11952.

\bibitem{owlvit}
Matthias Minderer, Alexey Gritsenko, Austin Stone, Maxim Neumann, Dirk Weissenborn, Alexey Dosovitskiy, Aravindh Mahendran, Anurag Arnab, Mostafa Dehghani, Zhuoran Shen, Xiao Wang, Xiaohua Zhai, Thomas Kipf, and Neil Houlsby,
\newblock ``Simple open-vocabulary object detection with vision transformers,''
\newblock {\em ECCV}, 2022.

\bibitem{metaclip}
Hu~Xu, Saining Xie, Xiaoqing~Ellen Tan, Po{-}Yao Huang, Russell Howes, Vasu Sharma, Shang{-}Wen Li, Gargi Ghosh, Luke Zettlemoyer, and Christoph Feichtenhofer,
\newblock ``Demystifying {CLIP} data,''
\newblock in {\em {ICLR}}, 2024.

\bibitem{instructblip}
Wenliang Dai, Junnan Li, Dongxu Li, Anthony Meng~Huat Tiong, Junqi Zhao, Weisheng Wang, Boyang Li, Pascale Fung, and Steven C.~H. Hoi,
\newblock ``Instructblip: Towards general-purpose vision-language models with instruction tuning,''
\newblock in {\em NeurIPS}, 2023.

\bibitem{filip}
Lewei Yao, Runhui Huang, Lu~Hou, Guansong Lu, Minzhe Niu, Hang Xu, Xiaodan Liang, Zhenguo Li, Xin Jiang, and Chunjing Xu,
\newblock ``{FILIP:} fine-grained interactive language-image pre-training,''
\newblock in {\em {ICLR}}, 2022.

\end{thebibliography}

\end{document}